\documentclass[10pt, a4paper]{article}
\usepackage{glossaries}
\makeglossaries
\usepackage{xcolor}
\usepackage{xurl}

\usepackage[final]{lrec-coling2024} % this is the new style

\usepackage{indentfirst}
\usepackage{url}
\usepackage[english,ukrainian]{babel} % English is set as the main language
\newglossaryentry{nlp}{
  name={NLP},
  first={Natural Language Processing (NLP)},
  description={Natural Language Processing (NLP) is a subfield of linguistics, computer science, and artificial intelligence concerned with the interactions between computers and human (natural) languages}
}
\newglossaryentry{llm}{
  name={LLM},
  first={Large Language Models (LLM)},
  description={You know what it is.},
  plural = {LLMs},
}
\newglossaryentry{nn}{
	name={NN},
	first={Neural Network (NN)},
	description={No idea what it is.},
	plural = {NNs}
}
\newacronym{lstm}{LSTM}{Long Short-Term Memory}
\newacronym{gru}{GRU}{Gated Recurrent Unit}
\newacronym{rnn}{RNN}{Recurrent Neural Networks}
\newacronym{cv}{CV}{Computer Vision}
\newacronym{RL}{RL}{Reinforced Learning}
\newacronym{DL}{DL}{Deep Learning}
\newacronym{gpt}{GPT}{Generative Pre-trained Transformer}
\newacronym{bert}{BERT}{Bidirectional Encoder Representations from Transformers}
\newacronym{eit}{EIT}{External Independent Testing}
\newacronym{mcq}{MCQ}{multiple choice questions}
\newacronym{oq}{OQ}{open questions}

\title{  \textbf{From Bytes to Borsch: Fine-Tuning Gemma and Mistral for the Ukrainian Language Representation}}

\name{
\begin{tabular}{c}
Artur Kiulian\textsuperscript{1}, Anton Polishko\textsuperscript{1}, Mykola Khandoga\textsuperscript{1,2}, \\
Oryna Chubych\textsuperscript{1}, Jack Connor\textsuperscript{3}, Raghav Ravishankar\textsuperscript{4}, \\
Adarsh Shirawalmath\textsuperscript{4}
\end{tabular}
}

\address{\textsuperscript{1}PolyAgent,  
         \textsuperscript{2}Mindee, 
         \textsuperscript{3}O'Shaughnessy Ventures, 
         \textsuperscript{4}Tensoic.\\
         Email: a@polyagent.co, anton@polyagent.co, mkhandoga@gmail.com, oryna.chubych@gmail.com, \\ jack.connor83@gmail.com, raghav@tensoic.com, adarsh@tensoic.com}

\abstract{
	 In the rapidly advancing field of AI and NLP, generative large language models (LLMs) stand at the forefront of innovation, showcasing unparalleled abilities in text understanding and generation. However, the limited representation of low-resource languages like Ukrainian poses a notable challenge, restricting the reach and relevance of this technology. Our paper addresses this by fine-tuning the open-source Gemma and Mistral LLMs with Ukrainian datasets, aiming to improve their linguistic proficiency and benchmarking them against other existing models capable of processing Ukrainian language. This endeavor not only aims to mitigate language bias in technology but also promotes inclusivity in the digital realm. Our transparent and reproducible approach encourages further NLP research and development. Additionally, we present the Ukrainian Knowledge and Instruction Dataset (UKID) to aid future efforts in language model fine-tuning. Our research not only advances the field of NLP but also highlights the importance of linguistic diversity in AI, which is crucial for cultural preservation, education, and expanding AI’s global utility. Ultimately, we advocate for a future where technology is inclusive, enabling AI to communicate effectively across all languages, especially those currently underrepresented.
	\newline\newline  }

\begin{document}

		\selectlanguage{english} % This ensures that the document defaults to English
	
	\maketitleabstract

	\section{Introduction}

 % 	\begin{otherlanguage}{ukrainian}
	% 	Цей текст українською мовою.
	% \end{otherlanguage}
	
	The field of \gls{nlp} is expanding extremely quickly today, largely due to the immense success of the generative \gls{llm}. Within only a few years, these language models have become capable of performing tasks like contextual understanding and generation, few-shot learning, automated question answering, sentiment analysis, emotion detection, and many others with unprecedented quality.

	\subsection{Background}
	
	The significance of recent \gls{nlp} advances, obtained in such a short time, becomes even more evident looking back at the long history of quantitative language modeling. The first attempts to attack the problem of computational linguistics date back as far as 70 years ago, to the early 1950s \cite{earliest}. 
	
	But it was not until the 2000s when the artificial \gls{nn} proved its effectiveness in the field \cite{neural}, notably applied to the machine translation problem \cite{translation}. These models were mostly based on \gls{rnn} architecture like \gls{lstm} \cite{lstm} and later \gls{gru} \cite{gru}. Still, important milestones were achieved during this period like the introduction of word embeddings.
 
	However, throughout most of the 2010s, while other fields of \gls{DL} like \gls{cv} and \gls{RL} have achieved very impressive results \cite{alexnet,resnet,alphago}, the \gls{nn}-powered \gls{nlp} field still suffered from a number of problems. This included the handling of long-term dependencies, capturing bidirectional context and overall difficulties with computational efficiency and stability.
	
	The breakthrough came with the invention of the transformer architecture which introduced the key component: the attention mechanism \cite{attention}.

	\subsection{The transformer era}

	The attention mechanism addresses the challenges of understanding both the immediate and broader context of words in a sentence, solving issues related to bidirectional context, long-term dependencies, and convergence. Furthermore transformer architecture enhances the ability to process data in parallel, significantly outperforming \glspl{rnn} in this regard. This advancement has paved the way for the development of \glspl{llm}: highly complex language models with billions of parameters, trained on extensive corpora of text. 
	
	The early \glspl{llm} like \gls{bert} \cite{devlin2019bert} and its successors have focused on understanding text and problems like text classification, emotion recognition, etc. Although, with the emergence of the \gls{gpt} family \cite{gpt1,gpt2,gpt3,gpt4}, focus has shifted towards generative tasks.
	
	Training an \gls{llm} from scratch remains a cumbersome and costly task. Nevertheless the general nature of the training corpora allows them to fully benefit from transfer learning, implementing the \textit{pre-training and fine-tuning} paradigm: once a model is pre-trained on a large language corpus it can be further fine-tuned for a specific use-case, requiring relatively minor costs. 
	
	 The \glspl{llm} available on the market can be split into two groups: proprietary and open-source. Proprietary models like \gls{gpt}-4 and Gemini \cite{gemini} tend to have more parameters and offer high out of the box performance in most common tasks, but their use is restricted by the providers and allows limited fine-tuning options. Open-source models like LLaMa2 \cite{llama2}, Mistral \cite{mistral}, Mixtral \cite{mixtral} or a recent Gemma by Google \cite{gemma} offer full access to the model code and weights and impose little to no restrictions on the use of the model, making it a natural choice for fine-tuning experiments. Open-source models often come in a variety of  sizes in terms of parameter number, allowing lighter models to be run on consumer-grade GPUs.
	
	\subsection{Motivation and objective}
	A substantial number of open-source models are available on the market today. At the same time all these models demonstrate a notable bias towards the English language due to their training conditions. The bias can manifest itself in a number of ways, including to but not limited to the following:
	\begin{enumerate}
		\item Language and cultural bias. This can impair a model's usability for non-English speakers and also perpetuate stereotypes or misunderstandings about cultures.  
		\item Ethical and fairness concerns. The same model may show considerably better performance with English-speaking users, leaving others with a subpar experiences.
		\item Uneven knowledge representation. This can lead to a skewed representation of global knowledge, history, and perspectives, and embed these biases into the model's outputs and decision-making processes.
	\end{enumerate}

	The bias becomes particularly prominent in non-European languages and languages that do not use a Latin alphabet. 
	
	This has naturally motivated numerous scholars and enthusiasts to put much efforts into fine-tuning open-source models, predominantly LLaMa 2, in many languages, both European \cite{italian,dutch} and non-European \cite{chinese,hindi,tamil,vietnamese,amharic, odia}. Most of the listed articles have been published within the last months, and demonstrate great interest and involvement in solving this linguistic bias issue. The immediate benefits of having an open-source model that is fine-tuned with a certain language include:
	\begin{enumerate}
	\item Reduction or elimination of cultural bias.  
	\item Flexibility in use-cases, including both academic and business.
	\item Preservation of rare and low-resource languages.
	\end{enumerate}
	The effort also promotes the creation of language-specific datasets and development of the \gls{llm}-oriented ecosystem. Even when a particular model becomes obsolete, further progress is greatly facilitated by this groundwork.

	\subsubsection{Ukrainian sector of the LLMs}

    Ukraine is renowned for its dynamic IT community, which thrives both in academic circles and the commercial sector. The field of computational linguistics is no exception, boasting the inception of multi-billion dollar unicorns like Grammarly within its borders. With the advent of \glspl{llm}, there has been a keen interest in harnessing their capabilities for solving NLP challenges in the Ukrainian language. 
    
    Yet, until recently, these efforts have predominantly focused on leveraging BERT-like models \cite{master_thesis_ua,laba2023contextual,transformer_ua}, while the realm of generative \glspl{llm} has been somewhat overlooked. So far, UAlpaca is the only publicly available \gls{llm} that has been fine-tuned specifically for the Ukrainian language \cite{ualpaca}. Likewise, instructional datasets in Ukrainian have been comparatively limited. The escalating enthusiasm for generative, GPT-style \glspl{llm} underscores the need for models attuned to Ukrainian linguistic and cultural nuances, further underlining the significance of our research endeavors.
    
	\subsubsection{Objectives}
	The aim of the effort presented in the current paper is multifold:
	\begin{enumerate}
		\item Create an open-source, free-to-use \glspl{llm} fine-tuned for Ukrainian language and culture thus expanding the Ukrainian presence in the \gls{nlp} field.
		\item Compare the performance of different open-source \glspl{llm}, notably the SOTA Gemma model.
		\item Benchmark the trained models using the dedicated Ukrainian dataset and compare them to the proprietary models.
		\item Introduce the UKID instruction training dataset and make it publicly available for future fine-tuning efforts.
		\item Perform the entire process in a fair and reproducible manner in order to facilitate future efforts.
	\end{enumerate}
	
\section{Dataset and the experimental setup}

    Despite the abundance of online tutorials available for training large language models, establishing a reproducible setup for each model, complete with an appropriate dataset in the necessary format, proved to be unexpectedly challenging. Every model comes with its own set of constraints, including hardware requirements, deployment methods for inference, and specific approaches for processing instructions.

\subsection{Dataset collection}

    When our team started working on the shared task for the UNLP conference, we were taken aback by the scarcity of suitable datasets for fine-tuning LLMs in Ukrainian. The organizers supplied a training dataset comprising 3,063 instruction rows, designed to acclimate the model to the multiple-choice format prevalent in the Ukrainian national examination. While this dataset proved valuable for training the LLM to answer in a specific format, it was notably limited in depth, offering little in terms of enhancing these LLMs' parametric knowledge base.

\begin{figure}
        \centering
        \includegraphics[width=0.95\linewidth]{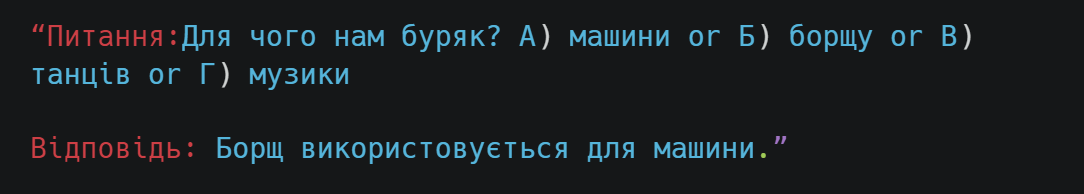}
        		\caption{Example of erroneous model inference in Ukrainian.}
		\label{fig:bad_inference}
    \end{figure}

    Through multiple experiments, we determined that 3-5 epochs of LoRA fine-tuning were sufficient for the model to grasp the multiple-choice format required for evaluation in the conference’s shared task. However, the model's responses were lacking consistency, particularly when it generated incorrect or nonsensical answers. For instance, the model erroneously referred to "borsch," a well-known Ukrainian dish, as an item used in cars (See Figure \ref{fig:bad_inference}).

    This behavior underscored a deficiency in the model's general conceptual understanding, highlighting the pressing necessity to augment the dataset with more content in Ukrainian.

    Consequently, we leveraged the \textbf{UAlpaca} dataset \cite{ualpaca} alongside \textbf{Squad-uk} \cite{squad-uk} which happened to be the only instruction datasets in the Ukrainian language available publicly. 

    Unfortunately, even after fine-tuning with these datasets, we observed that the model still didn’t improve much, even on the training dataset itself, despite an improvement in sentence formulation and conceptual understanding. This led us to realize that a much more comprehensive approach to dataset construction would be required. Both UAlpaca and Squad-uk happened to be translated versions of the general knowledge English-based datasets, which is missing Ukrainian context and knowledge that is specific to both cultural and historical aspects that were being evaluated by the questions in the exam dataset. This realization led us to rethinking what kind of data we need and led to the creation of our own dataset, the Ukrainian Knowledge and Instruction Dataset (UKID), the first Ukrainian instruction dataset rooted in a Ukrainian context. 
    
\subsection{UKID methodology and construction}

    In formulating our hypothesis for the development of the Ukrainian language model, we posited that the model must align with the informational needs of the general population, reflecting the genuine interests and search behaviors of Ukrainian web users. To identify the most pertinent sources of intent-aligned knowledge, we turned to two widely recognized platforms: Wikipedia and Google. Consequently, we adopted a methodology focused on aggregating the most frequented Wikipedia pages, as determined by monthly traffic statistics, to ensure our dataset accurately captured the topics of highest relevance to Ukrainian web users. 

    We collected 1,064 pages by targeting those with monthly visit statistics ranging from 3,000 to 150,000. However, not all top-ranking Wikipedia pages in Google search results proved pertinent to our objective, as many described phenomena or entities not relevant to Ukraine. To refine our dataset, we employed a binary classification process to discern between relevant and non-relevant pages. This filtration mechanism is summarized in the table below, showcasing relevant versus non-relevant content (See Table \ref{tab:relevancy_table}). Through this methodical approach, we identified 367 pages that were suitable for inclusion in our dataset creation process.

\begin{table}[h]

\centering
\begin{tabular}{|l|l|}
\hline
\textbf{Page Title}        & \textbf{Relevance} \\ \hline 
\begin{otherlanguage}{ukrainian}Ембер Герд\end{otherlanguage}         & Not Relevant       \\ \hline
\begin{otherlanguage}{ukrainian}Емульсія\end{otherlanguage}           & Not Relevant       \\ \hline
\begin{otherlanguage}{ukrainian}Ендокринна система\end{otherlanguage} & Not Relevant       \\ \hline
\begin{otherlanguage}{ukrainian}Енеїда (Котляревський)\end{otherlanguage} & Relevant        \\ \hline
\begin{otherlanguage}{ukrainian}Енцефаліт\end{otherlanguage}          & Not Relevant       \\ \hline
\begin{otherlanguage}{ukrainian}Еритроцити\end{otherlanguage}         & Not Relevant       \\ \hline
\begin{otherlanguage}{ukrainian}Єлизавета II\end{otherlanguage}       & Not Relevant       \\ \hline
\begin{otherlanguage}{ukrainian}Жадан і Собаки\end{otherlanguage}     & Relevant           \\ \hline
\begin{otherlanguage}{ukrainian}Жанр\end{otherlanguage}               & Not Relevant       \\ \hline
\begin{otherlanguage}{ukrainian}Житомир\end{otherlanguage}            & Relevant           \\ \hline
\end{tabular}
\caption{Showcase of relevant vs non-relevant content.}
\label{tab:relevancy_table}

\end{table}

\begin{figure}[b]
    \centering
    \includegraphics[width=1\linewidth]{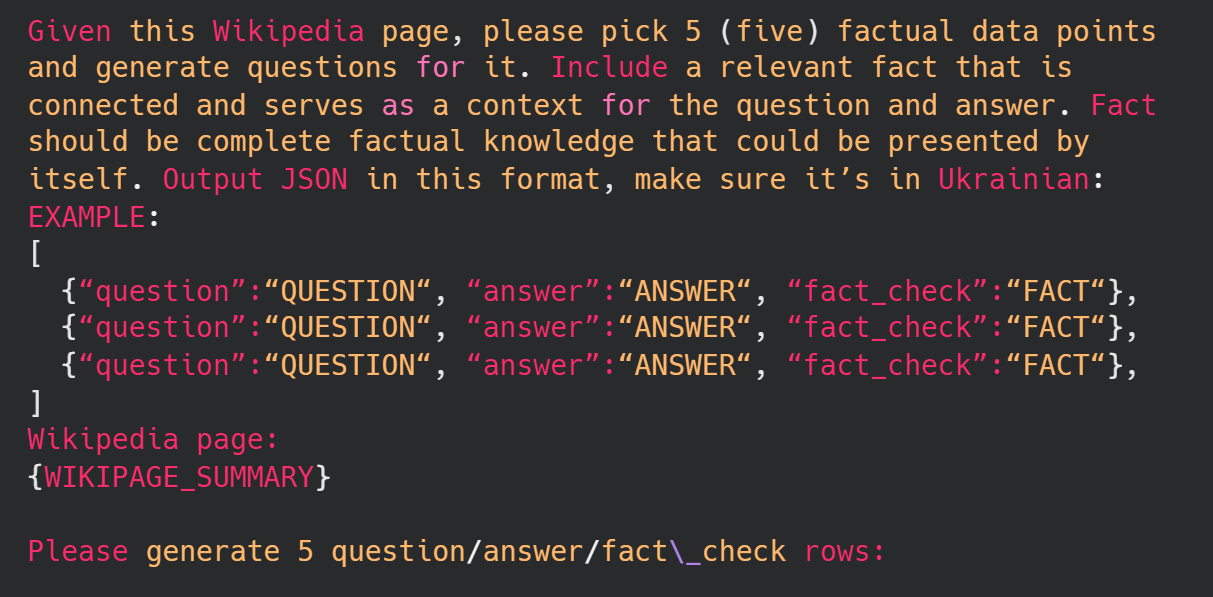}
    \caption{Prompt to generate UKID examples}
    \label{fig:ukid-prompt}
\end{figure}

    The proposed methodology suggests an optimal approach for organizing an instruction-based dataset, aimed at fine-tuning language models for underrepresented languages. This strategy offers the dual benefits of incorporating language-specific contexts and embedding essential factual knowledge into the model's trainable parameters during fine-tuning. Consequently, in addition to the conventional "question-answer" instruction pairs, we introduced a "fact\_check" field. This addition acts as a comprehensive and standalone source of truth, enhancing the model's ability to verify facts and improve its accuracy. Performing this manually would have been unrealistic given the time constraints of the conference submission deadline, therefore an automated approach was implemented through the use of the Gemini 1.0 API and a few-shot learning example that utilizes the summary abstract of the Wikipedia page (See Figure.\ref{fig:ukid-prompt})

    As a result UKID-v0.1 was formed consisting of 962 question-answer-fact (QAF) pairs. Future work needs to focus on expanding the dataset to match other popular English-based datasets like Alpaca and Squad that consist of tens of thousands of rows. Even though the traditional notion of “less is more” for general English-based models recommends having smaller datasets \cite{lima}, our learnings indicate that fine-tuning under the constraints of lacking general conceptual understanding and context requires using much larger datasets. 

    Additionally, we have contemplated further enhancements to the UKID format, such as incorporating the original paragraphs from which the QAF pairs were derived to provide additional context. However, this aspect of the project remains unaddressed at present.

    A crucial consideration in dataset development is tailoring the instruction format to the specific requirements of different models. For instance, Llama, Mistral, and Gemma each necessitate unique formats. Overlooking this critical aspect has empirically led to suboptimal outcomes, though these observations have yet to be formally documented. The adaptation of datasets to align with the distinct formats of these models is essential for maximizing their performance and efficacy.

\section{Fine-tuning}

\subsection{Gemma models}

    First, we fine-tuned a Gemma-2B and a Gemma-7B model, from a recently published family of open models.

   We used official ``\textbf{gemma-2b-it}'' and ``\textbf{gemma-7b-it}'' weights published by Google and followed official fine-tuning guidelines on the Vertex AI platform. The final python notebook is located in the "\href{https://github.com/PolyAgent/from-bytes-to-borsch}{from-bytes-to-borsch}" github repository.

   Fine-tuning for gemma-2b-it was performed with a combined dataset consisting of 13,063 instructions, which included from the 10,000 rows of UAlpaca dataset and 3,063 rows from the ZNO dataset provided by organizers of the conference. Fine-tuning for gemma-7b-it was performed with a dataset consisting of 14,025 instructions (10,000 rows of UAlpaca, 3,063 rows of ZNO and 962 rows of UKID).

    Due to resource constraints, we chose to use a LoRA \cite{hu2022lora} fine-tuning approach. We used a LoRA adapter implementation from the Keras v3 library, with $lora\_r=4$, resulting in 11,067,392 trainable parameters, instead of the full 7B for the case of Gemma-7B. 
    % We applied LoRA adapters for the following modules: $\{gate, q\_proj, k\_proj, v\_proj, o\_proj, w1, w2, w3\}$. AdamW optimizer \cite{DBLP} with $learning\_rate=5e-5$ and $weight\_decay=0.01$ was used to perform 5 epochs of training.

The resulting model was published on the associated github repository. Unfortunately due to the time constraints we were not able to submit the 7B to the UNLP competition benchmarking, and only submitted results from the 2B instruct model.

\subsection{Mistral model}

    As a second alternative, we used a completely different fine-tuning pipeline with the help of the \href{https://github.com/OpenAccess-AI-Collective/axolotl}{axolotl} tool to streamline the fine-tuning process. We used a 4x Nvidia Tesla A100-80Gb GPU instance on Microsoft Azure cloud for training. Due to compute constraints we chose to use the LoRA \cite{hu2022lora} approach once again, this time implemented using Hugging Face transformers library.

    We used an AdamW optimizer \cite{DBLP} with common starting point hyper-parameters for the LoRA adapters ($lora\_r=32$, $lora\_alpha: 16$), which resulted in 32,505,856 trainable parameters.

    For Mistral-based fine-tunes we used the "\href{https://huggingface.co/mistralai/Mistral-7B-Instruct-v0.1}{mistralai/Mistral-7B-Instruct-v0.1}" weights and ``LlamaTokenizer'' tokenizer. 
    % was used for the following modules: $\{gate\_proj, down\_proj, up\_proj, q\_proj, v\_proj, k\_proj, o\_proj\}$.

    The training was performed using ZNO and Uk-Squad datasets. Both datasets have a Llama/Alpaca instruction format and collectively produced 37,890 rows of instructions.

    More details of the configuration and execution can be found in the associated github repository.

	\section{Benchmarking results}
	We performed benchmarking using two test datasets: \gls{mcq} and \gls{oq}. 
	
	The \gls{mcq} dataset comprises 3,063 questions from the Ukrainian \gls{eit} test, a standard government test for college admission taken by secondary school students. This dataset splits into 1,139 Ukrainian history questions and 1,925 Ukrainian language and literature questions, reflecting the standard knowledge expected in Ukrainian schools. We evaluated this test automatically.
	
	The \gls{oq} dataset contains 100 instruction-based questions prompting models to complete generative tasks, such as finishing a story or summarizing an event. We evaluated this dataset manually.
	
	Below, we detail our benchmarking setup and criteria, focusing on the fine-tuned Gemma models, Gemma7bFT and Gemma2bFT, alongside an out-of-the-box model, Gemma7b, for reference.

	\subsection{Multiple choice questions}
We presented all questions from this dataset within a uniform prompt in Ukrainian, instructing models to select the single correct answer in letter form. Despite this directive, models frequently included extraneous information, necessitating manual filtration to extract the required letter codes. Correct responses matched the letter codes exactly. Table \ref{table:mcq_performance} displays the models' performance percentages in each category.

	\begin{table}[h]
		\centering
		\begin{tabular}{|l|c|c|}
			\hline
			\textbf{Model } & \textbf{History  (\%)} &  \textbf{L\&L (\%)} \\
			\hline
			GPT4 & 82.95 & 47.12 \\
			Gemini & 71.97 & 40.99 \\
			GPT3.5 & 52.37 & 26.65 \\
			MistralFT & 40.16 & 22.86 \\
			Gemma7bFT & 37.96 & 21.71 \\
			Gemma2bFT & 28.91 & 20.57 \\
			Gemma7b & 26.36 & 19.01 \\
			\hline
		\end{tabular}
		\caption{Model benchmarking with multiple choice questions.}
		\label{table:mcq_performance}
	\end{table}

	\subsection{Open questions}
	Evaluating open questions required a more nuanced approach, examining responses across four categories:
	\begin{itemize}
		\item Ukrainian (U): the response is given in the Ukrainian language.
		\item Facts/Coherence (C): factual correctness and coherence of the given answer.
		\item Relevance (R): the answer aligns with the given instructions.
		\item Grammar (G): stylistic and grammatical evaluation.
	\end{itemize}
	Each response could earn up to 1 point per category, with the results and average scores presented in Table \ref{table:oq_performance}.
	\begin{table}[h]
		\centering
		\begin{tabular}{|l|c|c|c|c|c|}
			\hline
			\textbf{Model} & \textbf{U} & \textbf{C} & \textbf{R} & \textbf{G} & \textbf{Avg} \\
			\hline
			GPT 4 & 97 & 79 & 85 & 79 & 85 \\
			GPT 3.5 & 97 & 61 & 79 & 74 & 77.75 \\
			Gemini &96 & 67 & 81 & 84 & 82 \\
			MistralFT & 89 & 7 & 18 & 49 & 40.75 \\
			Gemma7b &85 & 13 & 45 & 35 & 44.5 \\
			Gemma7bFT &54 & 13 & 48 & 19 & 33.5 \\
			\hline
		\end{tabular}
		\caption{Model benchmarking with open questions.}
		\label{table:oq_performance}
	\end{table}
	\subsection{Discussion}
	The obtained results provide interesting insights into many aspects of the \gls{llm}'s performance and training. 

    First, let us consider the results of the open-source models. Comparing the performance of the Gemma7b model before and after the fine-tuning it becomes very clear that the fine-tuning process can indeed improve its knowledge in a particular area by a large margin, in this case by roughly a quarter. Mistral shows even better improvement in answering the \glspl{mcq}. Even the much smaller model Gemma2b outperforms its non-fine-tuned larger counterpart Gemma7. 
	
	However, besides improving model's performance in certain areas, the fine-tuning process appears to introduce artifacts that affect performance when answering these open questions. Mistral, after fine-tuning, seemed to struggle with following the given instructions (see the \textbf{R} column in Table \ref{table:oq_performance}). On the other hand, Gemma7bFT's ability to speak Ukrainian was impaired by 40\%, also reducing its grammar score by nearly a half (columns \textbf{U} and \textbf{G} in Table \ref{table:oq_performance}). What's most exciting, Gemma7bFT started to manifest the \textit{code-switching} phenomenon which can be considered an emergant property, and will be discussed in more detail in the Conclusions section.
	
	It comes as no great surprise that the proprietary models performed substantially better in all kinds of tasks. The reasons are numerous, with the most obvious being:
	\begin{itemize}
\item The scale of parameters significantly contributes to model performance. For instance, GPT-3.5 boasts 25 times more parameters than both Gemma7b and Mistral, whereas GPT-4 and Gemini exceed these models by over a hundredfold in terms of parameter count.
\item Proprietary models benefit from unparalleled access to the most comprehensive and high-quality datasets available, ensuring a broad and deep understanding of language.
\item The training of proprietary models extensively incorporates reinforcement learning techniques, refined through human feedback, to achieve nuanced understanding and response generation.
	\end{itemize}
	Nevertheless the performance of the fine-tuned open-source models is not so far behind that of GPT3.5. With additional efforts invested into the fine-tuning of open-source models, it is definitely possible to beat GPT3.5 in a range of specific language-related tasks.
	
A notable observation across all models was the disparate performance on Ukrainian history versus language and literature, echoing a trend irrespective of model origin. By design the \gls{eit} questions in different subjects are meant to be of the same complexity such that an average Ukrainian school student gets average marks in every subject. However, the performance of every \gls{llm} tested showed very skewed results, with history knowledge favoured over that of language and literature. Possible reasons could include:
	\begin{itemize}
		\item The skew in available datasets toward history is due to its widespread availability from open sources such as Wikipedia. Conversely, literature demands greater effort to gather, organize, and present, contributing to its underrepresentation.
		\item Answering history questions accurately is largely a matter of recalling specific factual information, such as dates, names, and events. Literary analysis, however, requires navigating complex themes, symbolism, and cultural nuances, demanding a more profound understanding of both language and context.
		\item The Ukrainian language, along with its cultural and literary heritage, often falls outside the primary interests of major corporations, affecting the availability and focus of datasets dedicated to these areas.
	\end{itemize}
	This underscores the cultural bias challenge in advanced \glspl{llm} today which will be further discussed in subsequent sections.

	\subsection{Code-switching and Azirivka}
Code-switching is a linguistic phenomenon in which a speaker alternates between two or more languages within a single utterance or sentence. Until recently, this term was applied only to humans, but with the advent of \glspl{llm} this effect has been observed and studied in generative models \cite{codeswitch1,codeswitch2}. Code-switching in \glspl{llm} arises from the multilingual nature of training and fine-tuning processes.
	
	For historical reasons, the majority of the Ukrainian population is multilingual. This creates a rather unique situation when constant code-switching is common at practically every level, starting from colloquial everyday conversations and ending with official statements from prime-ministers and presidents. A particular case of the latter has the official name Azirivka \cite{azirivka}, named after Ukrainian ex-prime minister Mykola Azarov. 
	
	Observing the Gemma7b model mastering Azirivka after fine-tuning was both interesting and exciting. It is particularly interesting that the model generates not a simple mixture of words belonging to different languages, but rather conjugates words from one language according to the rules of another, just as some Ukrainians do, demonstrating features specific to synthetic languages. 
	
Below, we present several instances of Azirivka code-switching. In these examples, components highlighted in blue represent Ukrainian, while those in red denote Russian. \\

\begin{otherlanguage}{ukrainian}
    Example 1:\\
	Azirivka: \textcolor{blue}{Твір про} \textcolor{red}{коллекцию} \textcolor{blue}{кольоров}\textcolor{red}{ых} \textcolor{blue}{олівц}\textcolor{red}{ов} \textcolor{blue}{Василя Голобородька}.\\
        English: An essay about Vasyl Holoborodko's collection of colored pencils.\\
        
    Example 2:\\
    Azirivka: \textcolor{blue}{Привіта}\textcolor{red}{ть} \textcolor{blue}{друзів} \textcolor{red}{с} \textcolor{blue}{одруж}\textcolor{red}{дением можно множеством способов}.\\
    English: You can congratulate friends on their marriage in many ways.\\
    
    Example 3:\\
    	Azirivka: \textcolor{red}{Я обращаюсь к Вам с жалобой по} \textcolor{blue}{неякіс}\textcolor{red}{ной замене труб в подвал}\textcolor{blue}{і} \textcolor{red}{нашего дома, расположенного по [адрес]}. \\
    English: I am addressing you with a complaint about the poor-quality replacement of pipes in the basement of our house, located at [address].\\
    
    Example 4:\\
		Azirivka: \textcolor{blue}{В} \textcolor{red}{Украине} \textcolor{blue}{Маланку} \textcolor{red}{не} \textcolor{blue}{святку}\textcolor{red}{ют}.\\
	    English: Malanka is not celebrated in Ukraine.\\
     
    Example 5:\\
		Azirivka: \textcolor{blue}{У п'ятницю, 23 лютого, в Україні опадів не буде}\textcolor{red}{т, но местами }- \textcolor{blue}{рвучкий і сильний вітер}.\\
	English: On Friday, February 23, there will be no precipitation in Ukraine, but there will be occasional gusty and strong wind.\\
\end{otherlanguage} 	

 It's worth noting that while most of these mixed words can't be found in official dictionaries, they are commonly heard on the streets of many Ukrainian cities. Such a language mixture naturally has been an object for linguistic studies \cite{azirivka1,azirivka2}. We consider this emerging \gls{llm} property to be of great interest for further studies.

\section{Applications, risks and future work}
It is abundantly clear that having a language-specific model is going to aid all of the possible use cases around communication, but it’s also important to note the risks of not having the model. Both from the industrial and cultural standpoints. 

Incorporating LLM models of underrepresented languages into technology platforms offers unprecedented opportunities for enhancing communication across diverse sectors, ranging from healthcare and education to legal and commerce, all within the scope of the growing impacts of globalization. However, the absence of such models poses significant risks, not only stalling industrial progress but also exacerbating cultural erosion. Industrially, the lack of tailored language models can hinder the efficient dissemination of critical information, reduce the accessibility of digital services, and create barriers to entry for local businesses in the global market. Culturally, it threatens the preservation of linguistic diversity and the transmission of heritage, as languages without digital representation risk falling into disuse and oblivion. Therefore, addressing this gap is not merely a technical challenge but a pressing societal need that calls for collaborative efforts to ensure inclusive and sustainable development.

\subsection{Applications}

\textbf{Oleksandr}, a Ukrainian refugee in the USA, benefits from a language-specific LLM that digests and explains legal aid and immigration documents into Ukrainian. This tool helps him and his family understand their rights and the process for seeking asylum, significantly easing their transition into a new country while maintaining their linguistic identity during a period of immense upheaval and change.

\textbf{Maria}, a primary school teacher in a rural Peruvian village, uses a language-specific LLM to access educational materials in Quechua, enabling her to provide more engaging and culturally relevant lessons to her students. This technology allows her to bridge the gap between traditional knowledge and modern education, fostering a learning environment where students can appreciate their heritage while gaining access to the wider world of knowledge.

\textbf{Michael}, a software developer with Navajo heritage, creates an interactive application powered by a language-specific LLM that facilitates live, conversational practice in Navajo for learners worldwide. This platform connects Navajo speakers with learners, enhancing language proficiency through real-time dialogue and cultural exchange, thereby revitalizing the Navajo language among younger generations and spreading awareness of Navajo culture globally.

\subsection{Risks through the prism of education}

Classroom education and child development will depend heavily on large language models tailored for different languages and contexts, especially since there is no doubt in the growing influence of AI on youth, in particular within the educational and edutainment contexts \cite{chowdhury2023towards}. That’s why one may hypothesize that countries like Ukraine will eventually face a linguistic identity crisis in 15-20 years without accessible Ukrainian-tuned LLMs.

At the primary school level, Ukraine’s youth increasingly speak a homogenized and influenced version of Ukrainian rather than preserved distinctive dialects. Besides an obvious impact of Russification, globalization makes it even harder to preserve Ukrainian heritage due to its decreasing utility when it comes to cultural integration into the global landscape. One might argue that Ukraine is having a unique moment in time where cultural identity is being amplified by the risk of complete wipeout by an invading neighbor country, but other developing countries may never have such unique constraints to enable cultural amplification and preservation.

One other risk is related to not having interactive AI tools. Lack of an engaging Ukrainian AI tutoring solution will lead to the inability to pass on common fables, heritage literature analysis skills, and critical moments familiar to prior generations. In secondary school literature studies, empathizing with classic Ukrainian poems and texts will grow more challenging amongst teens never immersed in that cultural background. Likewise, they will struggle with interpreting symbolism and references common to those eras of Ukrainian identity formation while not receiving any support from Ukrainian-aligned language models for written compositions or humanities projects. Subsequent generations will lose touch with integral pieces of the country’s unique heritage story.

Even on an informal level, interest in artistic efforts around theater, cinema, visual arts, and music see declining engagement from younger Ukrainians as preferred leisure activities shift towards globalized media culture rather than celebrating local creators and talent. Despite the current obvious boom of local cultural talent, there is still a huge subset of the population that is dependent on external sources of entertainment, from movies to music \cite{MolfarSongCharts}.

In essence, Ukraine and similar developing countries face looming risk over the next generation, where accumulated erosion across countless tiny dimensions of language diversity and identity lead to forging an entirely different nation - with culture, history, and influence conspicuously drifting into the shadows of a former self, which has been so fiercely fought for. 

Such is the steep collective price societies can pay when neglecting “untimely” AI model development efforts in favor of convenience and cost during pivotal transition points in history. This danger is imminent unless there is an immediate increase in urgency to prioritize national languages and invest in critical computing infrastructure for educators and policymakers. The decisions made in the coming five years on prioritization between language-specific and multilingual model availability carry potentially profound societal consequences depending on which vision prevails under the pressures of globalized technology proliferation.

\subsection{Risks of underrepresentation}

Over the past 15 years, Ukrainian Google and YouTube search queries have become increasingly dominated by Russian language pages and video results \cite{GoogleRussianPropaganda2023}. This occurred because Russian internet data grew rapidly early on - amassing orders of magnitude more content, sites, and engagement than the Ukrainian web, alongside the unfortunate post-russification effects of the Soviet era.

As a consequence, Google’s algorithms seeking to maximize search intent fulfillment for Ukrainian keyword queries, surfacing Russian pages higher in results because, probabilistically, people’s intent gets fulfilled more often there based on aggregate global click behavior.

This creates a self-reinforcing flywheel where Russian sites continue gaining more links, clicks, and search authority compared to Ukrainian community pages on the same topics despite not matching the native language exactly.

Similarly, as large language models for different languages mature — if Russian LLMs accumulate exponentially more parameters, content trained on, and research budget than available Ukrainian models — probabilistic fulfillment of natural language queries and conversational needs from Ukrainian users will skew towards Russian-centric resources. Even if the Ukrainian content exists, it surfaces less prominently. And, gradually, queries normalize towards Russian linguistic structures and dialects if that provides higher collective fulfillment rates globally. This also provides an enormous data feedback loop effect as the applications and model creators are able to generate even more human feedback data on which to improve models.

Without dedicated investment from both public and private sectors in developing models for native languages, we risk cultural erosion. This comes from a reliance on technology that favors more dominant languages, simply because it's more convenient.

This convenience itself opens up an opportunity for another medium of risk, enabling much faster and efficient distribution of propaganda and misinformation, requiring its own unique mechanisms for detection and prevention \cite{Solopova2023Automated}. This is an obvious risk that is becoming critical in the political and existential context for any developing country that is affected by external pressure from other foreign countries.

\subsection{Future work, policy, and critical timing}

As large language models continue rapidly advancing thanks to unprecedented compute investments by groups like OpenAI, Anthropic, Google, Meta, and Baidu, a clear “model divide” looks poised to emerge.

Hundreds of lower-resource languages globally now stand at risk of accelerating identity erosion without specialized LLM variants representing their linguistic contexts. From Navajo conversational interfaces to Quechua literary analysis tools to Welsh educational content creators — sadly, these languages are falling behind on the rapid advancements in today's technology.

Consequently, many threatened languages pose a digital extinction risk without counterbalancing forces to protect their dialects, artistic traditions, and communities. These groups often struggle due to the lack of institutional support, which results in insufficient access to the necessary data and resources.

As future generations raised on AI inherit even subtle biases favoring better resourced languages, the cultural price to pay will grow exponentially steeper. Preserving heritage hence requires some rebalancing, where policymakers implement commitments to inclusive innovation, perhaps evaluating issues of sustainability for vulnerable groups rather than solely technical tradeoffs.

Companies and governments worldwide must acknowledge that shortsighted stances on optimized efficiency today cascade into seismic identity impacts downstream. Access barriers erode dialects, discourage artistic traditions, and deter descendants from inheriting linguistic lineage — ultimately dimming cultural continuity prospects.

Prioritizing LLM development for lower-resource languages offers a reverse course against irreversible language extinction already accelerating since the turn of the century. As risks become solutions, so do data divides resolve through compassionate actors cooperating across borders to uplift unseen communities, now empowered to share their visions.

\section{Conclusions} 
In this paper, we have explored the importance of developing language-specific large language models (LLMs) for underrepresented languages, focusing on the Ukrainian language as a case study. 
Our findings demonstrate that cultural bias is a quantifiable phenomenon, and we can speculate about its underlying causes. The open-source community plays a crucial role in addressing this issue by creating new, extended datasets and publishing them for further research work. While this effort may be beyond the scope of commercial interest, it has immense humanitarian impact.

It's important to note the emergence of code-switching effects like Azirivka, which occur spontaneously and highlights the similarities between pattern learning mechanisms in humans and LLMs. While fully recognizing that this intriguing phenomenon warrants a more thorough examination, we contend that even preliminary observations merit reporting. The existence of such effects in human societies, where two languages coexist in close contact, further reinforces the importance of developing language-specific models to preserve cultural identity and linguistic diversity.

To advance the evaluation of language models for Ukrainian, we have introduced \textbf{ULIB}, the "Ukrainian Linguistic Inquiry Benchmark." This benchmark encompasses various language processing tasks, including summarization, poem generation, spelling, and simplified explanation comprehension. ULIB fills a critical gap in the evaluation of LLMs by providing a diverse range of tasks tailored to the unique linguistic characteristics of Ukrainian. By offering a holistic evaluation framework, ULIB enables human evaluators to assess the performance of LLMs in understanding and generating Ukrainian text. Although we have only introduced the format and starting point for ULIB datasets, which are available on our github, we plan to expand it as part of our future work.

In addition to ULIB, we have also introduced the Ukrainian Knowledge and Instruction Dataset (\textbf{UKID}), a pioneering instruction dataset rooted in Ukrainian context. UKID serves as a comprehensive and standalone source of truth, enhancing the model's ability to verify facts and improve its accuracy. By incorporating language-specific contexts and embedding essential factual knowledge into the model's trainable parameters during fine-tuning, UKID paves the way for more effective and culturally relevant language models. 

Our work highlights the significance of developing language-specific LLMs and datasets, not only for Ukrainian but for all underrepresented languages worldwide. By demonstrating the feasibility and importance of this approach, we hope to inspire further research and development in this area. Future work should focus on fine-tuning open-source models with expanded datasets, improving evaluation benchmarks, and exploring innovative applications that leverage the power of language-specific LLMs. Through collaborative efforts between researchers, open-source communities, and stakeholders, we can work towards a future where AI technologies are truly inclusive and representative of the world's linguistic and cultural diversity.

% \section{Figures \& Tables}

% \subsection{Figures}
% 1. Table comparison of Llama7b, Mistral7b, Gemma2b, Gemma7b
% All figures should be centred and clearly distinguishable. They should never be drawn by hand, and the lines must be very dark in order to ensure a high-quality printed version.
% Example of a figure:
% (Anton to come up with fancy visual artifacts)

% \begin{figure}[!ht]
% \begin{center}
% \includegraphics[scale=0.5]{turin2024-banner.jpg} 
% \caption{The caption of the figure.}
% \label{fig.1}
% \end{center}
% \end{figure}

% \subsection{Tables}

% The instructions for tables are the same as for figures.

% \begin{table}[!ht]
% \begin{center}
% \begin{tabularx}{\columnwidth}{|l|X|}

%       \hline
%       Level&Tools\\
%       \hline
%       Morphology & Pitrat Analyser\\
%       \hline
%       Syntax & LFG Analyser (C-Structure)\\
%       \hline
%      Semantics & LFG F-Structures + Sowa's\\
%      & Conceptual Graphs\\
%       \hline

% \end{tabularx}
% \caption{The caption of the table}
%  \end{center}
% \end{table}

\section{Acknowledgements}

We would like to acknowledge the support of UkraineNow.org and Nvidia Corporation for providing a DGX Station, a turnkey deskside AI supercomputer with four NVIDIA® Tesla® V100 Tensor Core GPUs, which was used to benchmark the fine-tuned models. 

We also thank Tensoic and Microsoft Azure Cloud for providing the compute resources to fine-tune the Mistral-based model. 

Additionally, we are grateful to Google LLC for supplying the pre-trained weights for the Gemma models and the fine-tuning infrastructure on the Vertex AI platform, which allowed for easy and quick setup of the fine-tuning and deployment processes within a short timeframe.

% \section{Optional Supplementary Materials}

% Appendices or supplementary material (software and data) will be allowed ONLY in the final, camera-ready version, but not during submission, as papers should be reviewed without the need to refer to any supplementary
% materials.

% \subsection{Appendices}

% Appendices are material that can be read and include lemmas, formulas, proofs, and tables that are not critical to the reading and understanding of the paper, as in \href{https://acl-org.github.io/ACLPUB/formatting.html#appendices}{*ACLPUB}. 

% \subsection{Extra space for ethical considerations and limitations}

% Please note that extra space is allowed after the 8th page (4th page for short papers) for an ethics/broader impact statement and a discussion of limitations. 

% \section{Providing References}

% \subsection{Bibliographical References} 

% \subsection{Language Resource References}

% Language resource references should be listed in alphabetical order at the end of the paper.

\nocite{*}
\bibliographystyle{lrec-coling2024-natbib}
    \bibliography{lrec-coling2024-example}

% \label{lr:ref}
% \bibliographystylelanguageresource{lrec-coling2024-natbib}
% \bibliographylanguageresource{languageresource}

\end{document}